\documentclass[letterpaper]{article} 
\usepackage{aaai2026}  
\usepackage{times}  
\usepackage{helvet}  
\usepackage{courier}  
\usepackage[hyphens]{url}  
\usepackage{graphicx} 
\usepackage{natbib}  
\usepackage{caption} 
\usepackage{algorithm}
\usepackage{algorithmic}
\usepackage{amsmath}
\usepackage{amssymb}
\usepackage{mathtools}
\usepackage{amsthm}
\theoremstyle{plain}

\theoremstyle{definition}

\theoremstyle{remark}

\usepackage{multirow}
\usepackage{arydshln}
\usepackage{colortbl}
\definecolor{Gray}{gray}{0.95}

\usepackage{booktabs}

\usepackage{newfloat}
\usepackage{listings}
\DeclareCaptionStyle{ruled}{labelfont=normalfont,labelsep=colon,strut=off} 
\floatstyle{ruled}
\newfloat{listing}{tb}{lst}{}
\floatname{listing}{Listing}

\title{MoFu: Scale-Aware Modulation and Fourier Fusion for \\ Multi-Subject Video Generation}
\author{
Run Ling\textsuperscript{\rm 1,\rm 2}\equalcontrib, 
Ke Cao\textsuperscript{\rm 3}\equalcontrib, 
Jian Lu\textsuperscript{\rm 4}\equalcontrib, 
Ao Ma\textsuperscript{\rm 1}\thanks{Project leader.}\thanks{Corresponding author.},
Haowei Liu\textsuperscript{\rm 4},
Runze He\textsuperscript{\rm 5}, \\
Changwei Wang\textsuperscript{\rm 5},
Rongtao Xu\textsuperscript{\rm 5},
Yihua Shao\textsuperscript{\rm 5},
Zhanjie Zhang\textsuperscript{\rm 1},
Peng Wu\textsuperscript{\rm 6}, 
Guibing Guo\textsuperscript{\rm 2\ddag}, \\
Wei Feng\textsuperscript{\rm 1},
Zheng Zhang\textsuperscript{\rm 1}, 
Jingjing Lv\textsuperscript{\rm 1}, 
Junjie Shen\textsuperscript{\rm 1}, 
Ching Law\textsuperscript{\rm 1}, 
Xingwei Wang\textsuperscript{\rm 2}
}

\affiliations{
    \textsuperscript{\rm 1}JD.com, Inc. \\
    \textsuperscript{\rm 2}Northeastern University \\
    \textsuperscript{\rm 3}University of Science and Technology of China \\
    \textsuperscript{\rm 4}Chongqing University of Post and Telecommunications \\
    \textsuperscript{\rm 5}University of Chinese Academy of Sciences \\
    \textsuperscript{\rm 6}Northwestern Polytechnical University \\

    \{lingrun.1, maao.8\}@jd.com, guogb@swc.neu.edu.cn
}

\usepackage{bibentry}

\begin{document}

\maketitle


\begin{abstract}
Multi-subject video generation aims to synthesize videos from textual prompts and multiple reference images, ensuring that each subject preserves natural scale and visual fidelity. However, current methods face two challenges: scale inconsistency, where variations in subject size lead to unnatural generation, and permutation sensitivity, where the order of reference inputs causes subject distortion.
In this paper, we propose MoFu, a unified framework that tackles both challenges. For scale inconsistency, we introduce Scale-Aware Modulation (SMO), an LLM-guided module that extracts implicit scale cues from the prompt and modulates features to ensure consistent subject sizes. To address permutation sensitivity, we present a simple yet effective Fourier Fusion strategy that processes the frequency information of reference features via the Fast Fourier Transform to produce a unified representation. Besides, we design a Scale-Permutation Stability Loss to jointly encourage scale-consistent and permutation-invariant generation.
To further evaluate these challenges, we establish a dedicated benchmark with controlled variations in subject scale and reference permutation. Extensive experiments demonstrate that MoFu significantly outperforms existing methods in preserving natural scale, subject fidelity, and overall visual quality.
\end{abstract}

\section{Introduction}








\begin{figure}[t]   
    \centering
    \includegraphics[width=\linewidth,scale=1.00]{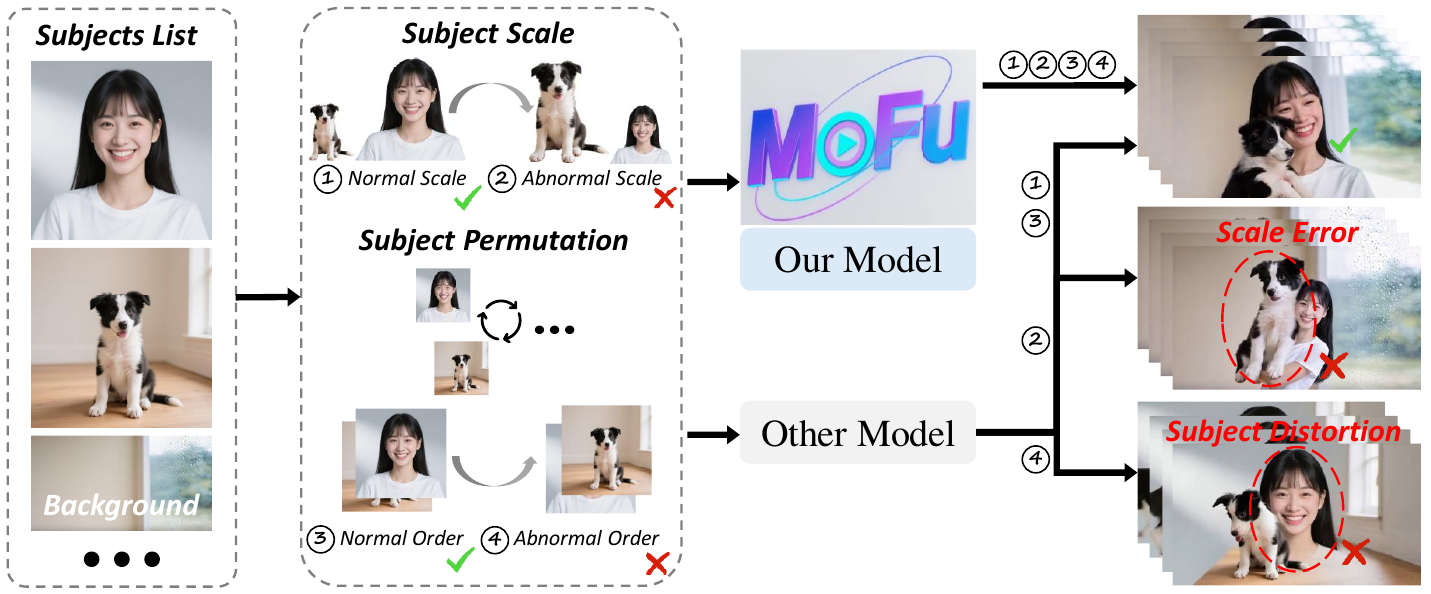}
    \caption{\label{fig:fig1}\textbf{Illustration of the challenges in multi-subject video generation.} Existing models often suffer from scale inconsistency and permutation sensitivity. Our MoFu framework addresses both issues, producing natural-scale and permutation-invariant videos.}
    \label{fig:dataset_pipeline}
\end{figure}

Multi-subject video generation aims to synthesize videos from textual prompts and multiple reference images, producing temporally consistent results where each subject preserves visual fidelity, appears at a natural scale, and remains unaffected by permutations of reference images.
Despite recent progress, multi-subject video generation faces two challenges. 
The first is \textbf{scale inconsistency}: reference images often vary significantly in subject scales due to different zoom levels, causing subjects in generated videos to appear unnaturally large or small.
Another is \textbf{permutation sensitivity}: existing methods process images sequentially, inserting them one by one along the frame or channel dimension. This leads to subject distortion, disappearance, or physically implausible interactions that depend on the input order. Moreover, as the number of references increases, computational cost rises sharply and efficiency declines, making sequential processing increasingly impractical.
Fig.~\ref{fig:fig1} illustrates these challenges, where existing models fail to maintain natural scales or steadily handle permutations of reference images.

Several recent works~\cite{phantom,skyreels} attempt to address scale inconsistency by fusing textual prompts into the reference image representation, leveraging spatial relationships implicitly encoded in the prompt (e.g., a girl is in the room with a dog in her arms). However, without an explicit mechanism to interpret and enforce natural scale, these methods produce subjects with unnatural scales that contradict the prompt, especially when the reference images themselves vary significantly in scale.
To address permutation sensitivity, some methods~\cite{customvideo,magref} concatenate multiple reference images onto a single blank canvas to form a unified visual input. While this avoids sequential processing, it introduces spatial biases (i.e., central or larger subjects may be prioritized) determined by subject placement in the composite image. Consequently, certain subjects may receive disproportionate attention or semantic weight, leading to distortion or disappearance when the spatial arrangement changes.


To address these challenges, we propose MoFu, a unified framework that tackles both scale inconsistency and permutation sensitivity in multi-subject video generation. 
For the former, we introduce Scale-Aware Modulation (SMO), an LLM-based module that extracts implicit scale relationships from the prompt and injects the condition into the model via modulation, ensuring subjects appear at natural scales.
For the latter, to fuse the reference images, we draw inspiration from a theorem in high-dimensional probability: ``the inner product between random vectors tends to zero, implying near-orthogonality~\cite{vershynin2018high}". Building on this insight, we propose a simple yet effective Fourier Fusion strategy.
Reference images are first segmented and encoded into fine-grained features, which are then transformed into the frequency domain using the Fast Fourier Transform (FFT). Their high- and low-frequency components are aggregated by direct summation, leveraging their near-orthogonality, and reconstructed via Inverse FFT (IFFT) to produce a permutation-invariant representation.
To further enforce these objectives, we introduce the Scale-Permutation Stability Loss, which jointly enforces scale consistency and permutation-invariant generation, improving subject fidelity and visual quality.

We propose a high-quality dataset, MoFu-1M, to train our model. Besides, to further evaluate the challenges of multi-subject video generation, we construct a dedicated benchmark, MoFu-Bench, that systematically evaluates models under varying subject scales and reference permutation. Extensive experiments show that MoFu consistently outperforms existing methods by achieving superior scale consistency and permutation-invariant generation, while preserving subject fidelity and overall visual quality.  

Overall, our key contributions are summarized as follows:  
\begin{itemize}
\item We propose MoFu, a unified framework that simultaneously addresses scale inconsistency and permutation sensitivity in multi-subject video generation.  
\item We design Scale-Aware Modulation, which dynamically adjusts feature representations using scale cues derived from the prompt, ensuring natural subject appearance despite scale variations in the references.  
\item We introduce a simple yet effective Fourier Fusion strategy that aggregates reference features in the frequency domain, leveraging the near-orthogonality property to produce a permutation-invariant representation.  
\item We build a high-quality dataset, MoFu-1M, for training, and establish a dedicated benchmark, MoFu-Bench, for assessing scale consistency and permutation-invariance, demonstrating that MoFu significantly outperforms SOTA baselines.
\end{itemize}

\section{Related Work}
\subsection{Video Generation Models}

Recent advances in content generation typically adopt Variational Autoencoders (VAEs)\cite{vae} to project visual data into compact latent spaces, enabling efficient large-scale generative pre-training. On top of these latent representations, modern video generation systems are predominantly built upon diffusion-based models\cite{wan2025,hunyuanvideo,ltx,opensora2}, auto-regressive frameworks~\cite{magi,nova,pavdm,he2025plangen}, as well as several high-capacity closed-source models~\cite{kling,hunyuanvideo,pika,vidu,sora,runway,seedance}. These approaches have substantially improved video realism, temporal coherence, and scalability.
Such progress has facilitated increasingly controllable generation paradigms, including text-to-video~\cite{wang2024qihoo,feng2024fancyvideo}, image-to-video~\cite{videomage}, motion-customized video generation~\cite{customttt,wang2025wisa,tora2}, and video style transfer~\cite{zhang2024artbank,ma2025lay2story,chiu2024styledit,lu2025uni,cao2025relactrl,ling2025ragar,bi2024using,wang2025learning,xu2025dropoutgs}. More recently, multi-subject video generation~\cite{phantom,skyreels,vace,magref,customvideo} has attracted growing attention, aiming to synthesize videos containing multiple entities with distinct appearances and identities. Compared to single-subject settings, this task introduces additional challenges in subject disentanglement, identity preservation, and compositional consistency, which remain difficult for existing methods, especially under complex reference conditions.

\subsection{Multi-Subject Video Generation}
Multi-subject video generation aims to synthesize videos containing multiple entities conditioned on prompts and reference images. Existing approaches tackle subject confusion through text–visual binding~\cite{conceptmaster,openset,fastcomposer}, LLM-guided layout reasoning~\cite{cinema,magiccomp}, or unified frameworks~\cite{vace}. However, two challenges remain: scale inconsistency and permutation sensitivity.
\textbf{Scale Inconsistency.}
Although prior works fuse prompt and reference cues to model spatial relations~\cite{phantom,skyreels}, they lack explicit mechanisms to preserve realistic subject scales, leading to unnatural size variations or contradictions with prompt semantics, especially when reference scales differ greatly.
\textbf{Permutation Sensitivity.}
To reduce sequential bias, some methods concatenate reference images~\cite{magref,customvideo} or apply cross-attention conditioning~\cite{weaver}, yet both introduce order-dependent artifacts or lose fine-grained identity cues. As a result, subject appearance often becomes distorted or inconsistent when the input order changes.




\section{Method}
This section presents the implementation details of MoFu.  
We first review the foundation of our framework, including the Diffusion Transformer (DiT) backbone and its modulation mechanism.  
Next, we describe the construction of a high-quality multi-subject dataset, MoFu-1M, and benchmark, MoFu-Bench.
Finally, we detail how Scale-Aware Modulation and Fourier Fusion are integrated into the video generation pipeline to explicitly address scale inconsistency and permutation sensitivity, and how the Scale-Permutation Stability Loss enforces these constraints during training.  

\begin{figure*}[htbp]
    \centering
    \includegraphics[width=0.95\textwidth]{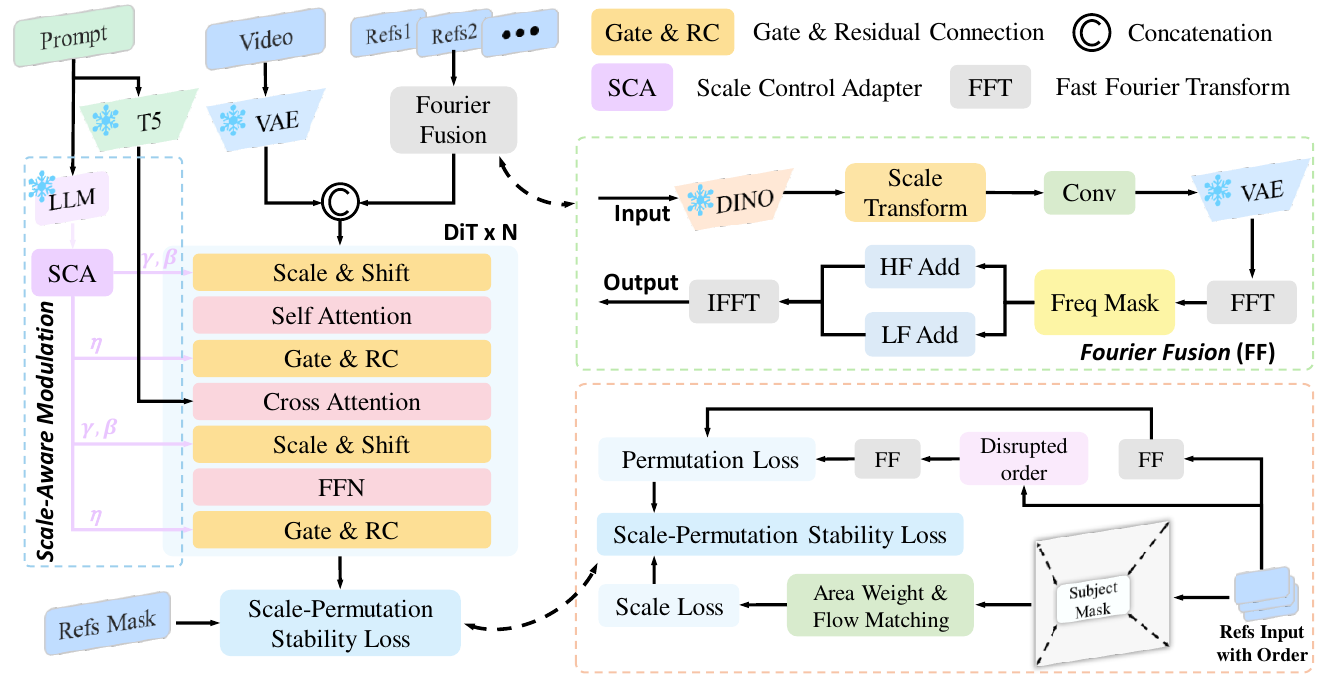}
    \caption{\label{fig:overview}\textbf{Overview of the MoFu framework.} MoFu integrates Scale-Aware Modulation (SMO), Fourier Fusion, and the Scale-Permutation Stability Loss (SPSL) into a DiT backbone. SMO extracts scale cues from the prompt via an LLM and adaptively modulates scale features to maintain natural subject scales. Fourier Fusion aggregates reference features in the frequency domain to form a permutation-invariant representation. Furthermore, SPSL jointly enforces scale consistency and permutation-invariant generation.}
    \label{fig:overview}
\end{figure*}
\subsection{Preliminary}

\subsubsection{Diffusion Transformer (DiT) and the Modulation}

DiT~\cite{dit} is a diffusion-based generative model that adopts a vanilla Transformer architecture, replacing the U-Net structures commonly used in latent diffusion models. This design enhances the ability to model long-range spatial dependencies and improves scalability.
A key strength of DiT lies in its modulation mechanism, which integrates external conditioning signals in a structured manner. Instead of directly concatenating or adding conditioning vectors, DiT applies adaptive scaling and shifting through normalization layers. Given a hidden representation $h$ and a modulation vector $m$, the modulation is defined as:
$$
\text{Mod}(h) = \gamma \cdot \text{LayerNorm}(h) + \beta,
$$
where the scaling and shifting parameters $\gamma$ and $\beta$ are derived from $m$ via lightweight MLPs. This mechanism enables the model to dynamically adjust internal representations according to conditioning inputs, effectively guiding the generative process with external conditions.

\begin{figure}[h]   
    \centering
    \includegraphics[width=\linewidth,scale=1.00]{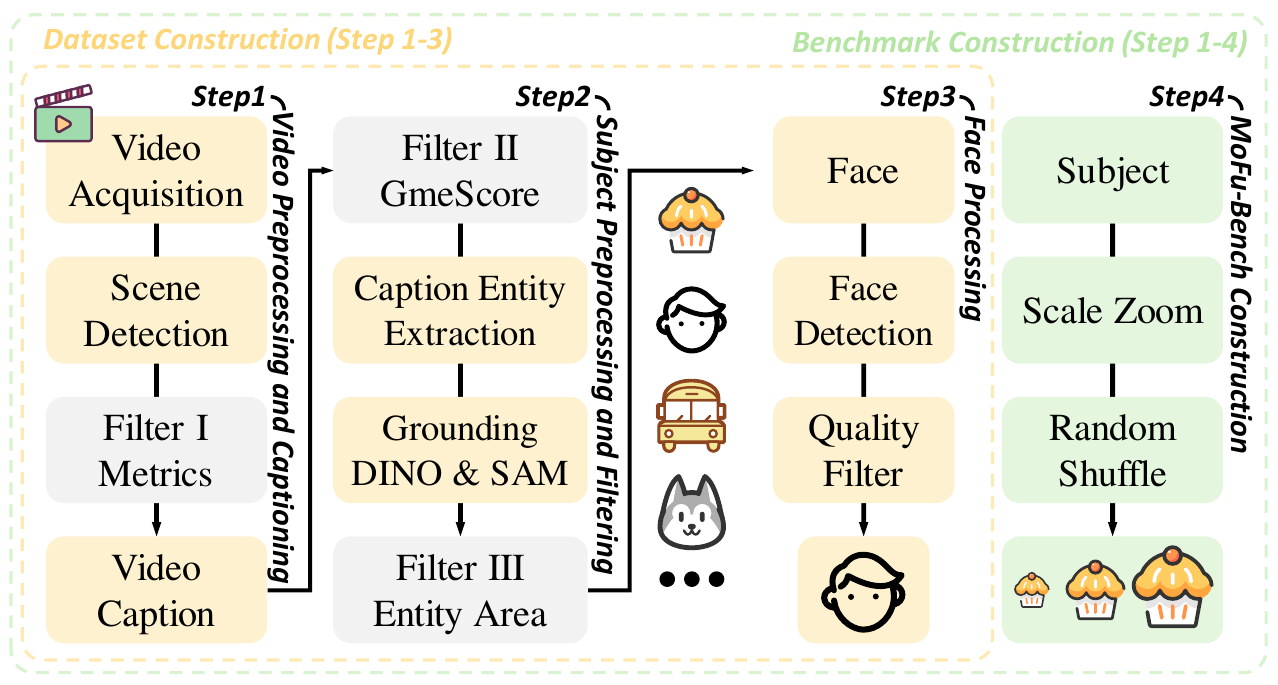}
    \caption{\label{fig:dataset_pipeline}\textbf{Construction pipeline of the dataset and benchmark.} We first perform video preprocessing and captioning, and multi-subject extraction and filtering. Multi-scale data for subjects and faces are then generated and refined, resulting in a dataset tailored for evaluating scale consistency and permutation-invariance in multi-subject video generation.}
    \label{fig:dataset_pipeline}
\end{figure}




\subsection{Dataset and Benchmark}\label{sec:dataset}
To train and evaluate MoFu, we construct a high-quality multi-subject video dataset, \textbf{MoFu-1M}, and a dedicated benchmark, \textbf{MoFu-Bench}, specifically designed to assess scale consistency and permutation invariance (see Fig.~\ref{fig:dataset_pipeline}).  

\begin{itemize}
\item \textbf{Video Preprocessing and Captioning.} Large-scale raw videos are segmented into coherent clips using scene detection~\cite{pyscenedetect}. Low-quality clips are filtered via aesthetic~\cite{aes} and motion metrics, and the remaining clips are automatically captioned with Qwen2.5-VL~\cite{qwen2.5vl}.  
\item \textbf{Subject Processing and Filtering.} We first filter out videos with poor text–video alignment using GmeScore. Next, we extract entities (e.g., people, animals, objects) from captions using LLM-based parsing~\cite{glm4} and localize them in video frames with Grounded-SAM~\cite{groundedsam}. Finally, we filter out videos where the extracted subjects are either too large or too small based on their relative area in the frames.  
\item \textbf{Face Processing.} Faces, which require higher fidelity, undergo additional detection and filtering~\cite{face} to preserve identity clarity, completing the final MoFu-1M dataset.  
\item \textbf{MoFu-Bench Construction.} To rigorously evaluate the model, we build MoFu-Bench by introducing controlled variations to the reference images. Specifically, we apply subject-centric zoom operations to create scale inconsistencies and randomly shuffle the order of references to simulate permutation changes. This benchmark directly tests a model's ability to maintain natural scale relationships and permutation-invariant generation under challenging conditions.
\end{itemize}

\subsection{MoFu framework}

As illustrated in Fig.~\ref{fig:overview}, our MoFu framework extends a DiT-based backbone with two core modules: Scale-Aware Modulation (SMO) for scale consistency and Fourier Fusion for permutation-invariant conditioning, while the Scale-Permutation Stability Loss (SPSL) jointly enforces these objectives during training.

\subsubsection{Scale-Aware Modulation}
The purpose of SMO is to explicitly capture scale relationships from the textual prompt and inject them into the generation process. 
We first encode the prompt $p$ using a frozen LLM~\cite{qwen2.5} to obtain a semantic embedding:
\begin{equation}
\mathbf{e}_p = \text{LLM}(p) \in \mathbb{R}^{d}.
\end{equation}
This embedding is then passed through a lightweight Scale Control Adapter (SCA), implemented as an MLP, to predict modulation parameters:
\begin{equation}
\gamma, \beta, \eta = f_{\text{SCA}}(\mathbf{e}_p),
\end{equation}
where $\gamma, \beta \in \mathbb{R}^{d}$ are scale and shift factors applied to normalized features, and $\eta \in \mathbb{R}^{d}$ is a gating factor to control residual connections.  

Given a DiT block feature map $\mathbf{F} \in \mathbb{R}^{B \times L \times d}$, SMO performs adaptive feature modulation:
\begin{equation}
\hat{\mathbf{F}} = \gamma \odot \mathbf{F} + \beta, \quad
\mathbf{F}_{out} = \mathbf{F} + \eta \cdot \text{Layer}(\hat{\mathbf{F}}),
\end{equation}
where $\odot$ denotes element-wise multiplication and $\text{Layer}$ represents either the MHA or FFN module.  
This modulation mechanism explicitly injects scale-aware cues into the features, thereby enforcing natural and consistent subject scales throughout the video generation process.
%



\begin{figure*}[h]
    \centering
    \includegraphics[width=\textwidth]{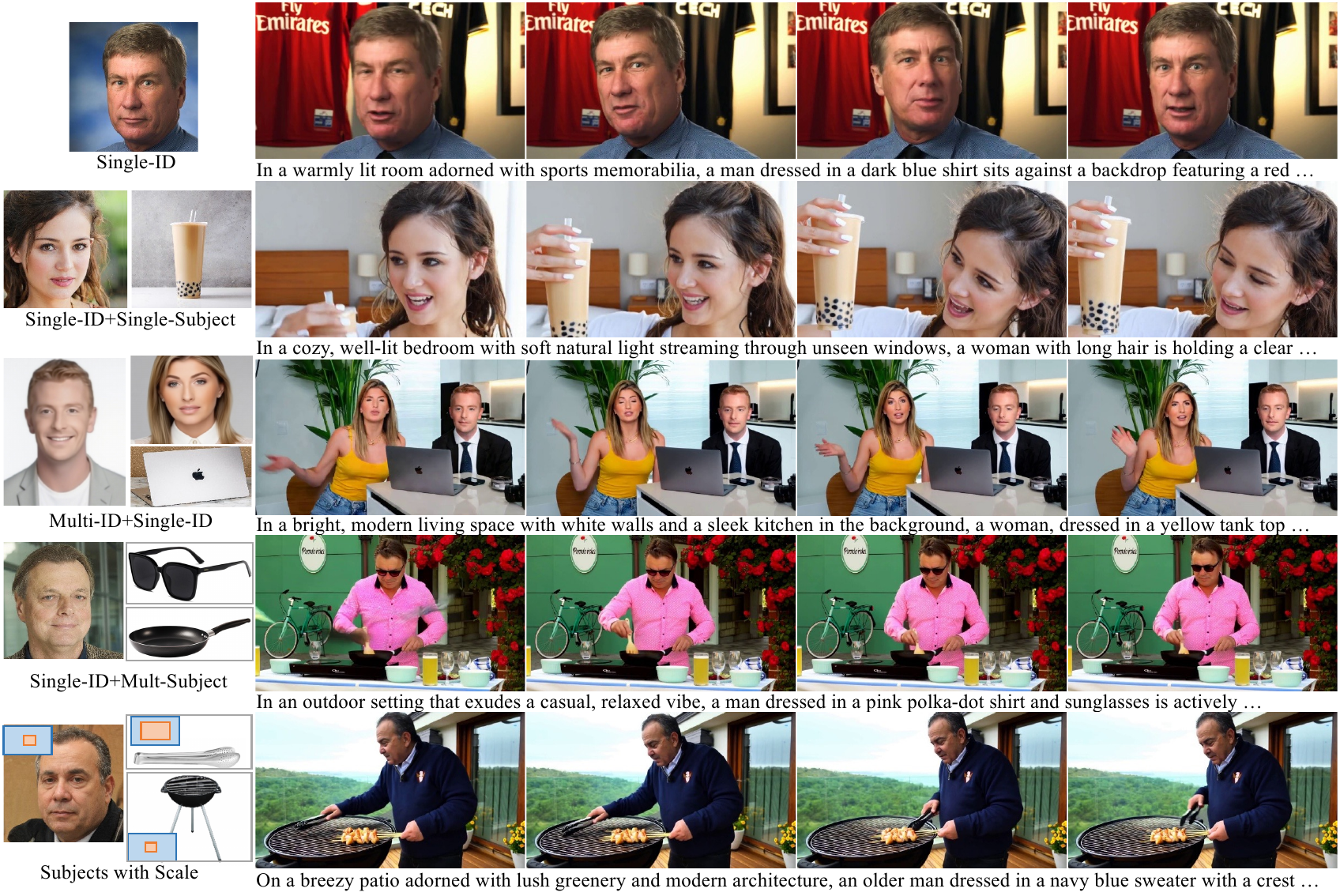}
    \caption{Qualitative evaluation results of our method on different cases. Our model consistently generates video that maintains natural scale and permutation-invariance while accurately following the input text prompt. In row 5, the small rectangles overlaid on the reference images represent the relative area occupied by each subject in the frame. Specifically, the blue regions denote the padded white background, while the red regions indicate the actual subject area.}
    \label{fig:qualitative}
\end{figure*}

\subsubsection{Fourier Fusion}

To address permutation sensitivity, we propose the Fourier Fusion strategy, which aggregates multiple reference images without introducing positional or order bias.
Given $N$ reference images $\{x_1, x_2, \ldots, x_N\}$, we first apply Grounded-SAM to segment each image and obtain subject masks, from which the subject regions are cropped and resized. Each processed reference is then passed through a $3 \times 3$ CNN encoder $\mathcal{E}$ to extract the corresponding feature maps:
\begin{equation}
\mathbf{F}_i = \mathcal{E}(x_i) \in \mathbb{R}^{d \times H \times W}.
\end{equation}

Each feature map is transformed into the frequency domain using FFT:
\begin{equation}
\mathcal{F}_i = \text{FFT}(\mathbf{F}_i).
\end{equation}
We then decompose $\mathcal{F}_i$ into high-frequency (HF) and low-frequency (LF) components using a frequency mask $M_{\text{freq}}$:
\begin{equation}
\mathcal{F}_i^{\text{HF}} = M_{\text{freq}} \odot \mathcal{F}_i, \quad
\mathcal{F}_i^{\text{LF}} = (1 - M_{\text{freq}}) \odot \mathcal{F}_i,
\end{equation}
where $M_{\text{freq}}$ is a binary mask derived from the radial frequency map that separates HF and LF regions.

Leveraging the near-orthogonality of high-dimensional vectors, we aggregate these components across all references by simple summation:
\begin{equation}
\mathcal{F}^{\text{HF}} = \sum_{i=1}^{N} \mathcal{F}_i^{\text{HF}}, \quad
\mathcal{F}^{\text{LF}} = \sum_{i=1}^{N} \mathcal{F}_i^{\text{LF}}.
\end{equation}
The fused frequency representation is reconstructed via IFFT:
\begin{equation}
\mathbf{F}_{\text{fused}} = \text{IFFT}(\mathcal{F}^{\text{HF}} + \mathcal{F}^{\text{LF}}).
\end{equation}
This permutation-invariant representation is then concatenated with the video features to condition the generation process.

\subsubsection{Scale-Permutation Stability Loss}

To jointly enforce scale consistency and permutation-invariant generation, we design the Scale-Permutation Stability Loss (SPSL). SPSL consists of two complementary terms: a scale loss, which encourages the model to respect the relative scale of each subject, and a permutation loss, which learns permutation-invariant generation and penalizes inconsistencies caused by changing the order of reference images.

\paragraph{Scale loss.}  
The scale loss leverages reference masks and their relative area ratios to adaptively re-weight the standard flow-matching loss.
Given predicted and target noise tensors $\epsilon_{\text{pred}}, \epsilon_{\text{true}} \in \mathbb{R}^{B \times C \times T \times H \times W}$, and their base mean-squared error (MSE) loss $\mathcal{L}_{\text{mse}}$, we compute normalized spatial weights from the reference masks:
\begin{equation}
    \mathbf{M} = 
    \sum_{r=1}^{R} w_{r} \cdot 
    \text{Resize}(m_{r}), \quad
    w_{r} = \frac{\exp(a_{r})}{\sum_{r'} \exp(a_{r'})},
\end{equation}
where $m_{r}$ is the mask of the $r$-th reference image for the sample, $a_{r}$ is its area ratio and $w_{r}$ is the normalized weight. The resized masks are combined using the area ratio weights.  
The final scale loss is:
\begin{equation}
    \mathcal{L}_{\text{scale}} = 
    \frac{\sum (\mathcal{L}_{\text{mse}} \odot \mathbf{M})}
         {\sum \mathbf{M} + \epsilon},
\end{equation}
where $\odot$ denotes element-wise multiplication and $\epsilon$ is a small constant to avoid division by zero. This weighting scheme places stronger emphasis on accurately reconstructing subjects with larger relative areas while ensuring all frames remain normalized.

\paragraph{Permutation loss.}  
To ensure that the change in the permutation of references does not affect the generation, we compute an MSE loss in the frequency domain across $P$ permutations of reference inputs:
\begin{equation}
\mathcal{L}_{\text{prem}} = \frac{1}{P} \sum_{p=1}^P \left\| F(\mathcal{R}_p) - F(\mathcal{R}_{\text{ref}}) \right\|_2^2,
\end{equation}
where $F(\cdot)$ denotes the Fourier Fusion output, $\mathcal{R}_p$ is a permutation of references.

\paragraph{Final objective.}  
The complete SPSL is a weighted combination of the above terms:
\begin{equation}
    \mathcal{L}_{\text{SPSL}} = 
    \mathcal{L}_{\text{scale}} + 
    \mathcal{L}_{\text{perm}}.
\end{equation}
SPSL provides explicit guidance for MoFu to maintain natural subject scales and robust conditioning regardless of the input reference permutation.
\begin{figure*}[t]
    \centering
    \includegraphics[width=\textwidth]{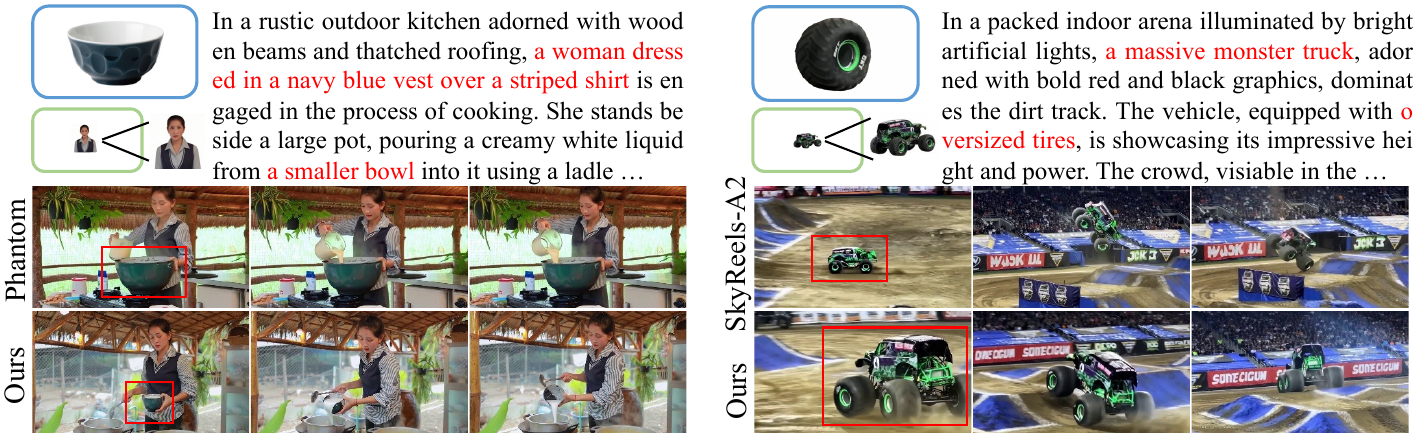}
\caption{Qualitative comparison on scale-inconsistent scenarios. 
In the left case, Phantom fails to maintain the correct relative scale of the bowl, while our approach produces a proportionally accurate result. 
Similarly, in the right case, the large monster truck is generated at a natural size relative to its environment.}
    \label{fig:compare_scale}
\end{figure*}
\begin{figure*}[h!]
    \centering
    \includegraphics[width=\textwidth]{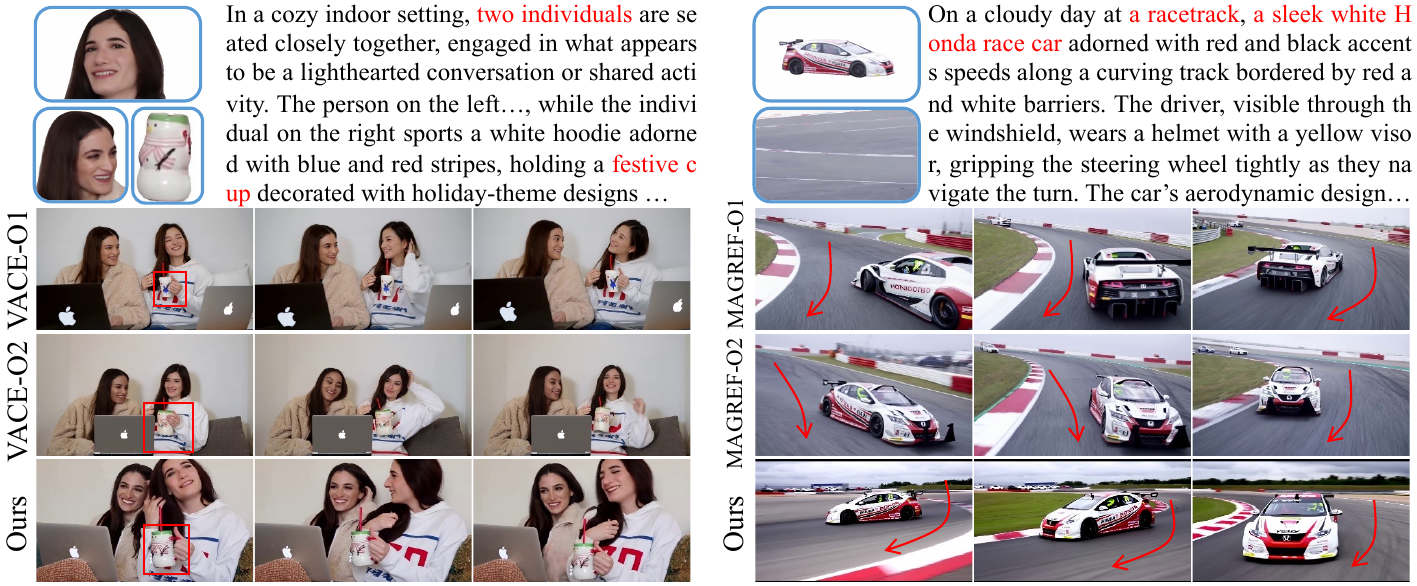}
    \caption{Qualitative comparison on permutation-variant scenarios.
    In the left case, other methods fail to maintain the correct representation of the cup when the input order changes, while our approach preserves it faithfully. 
    In the right case, changing the input order causes other methods to generate a race car moving backwards, violating physical plausibility, whereas our method consistently produces videos with correct orientation and motion.}
    \label{fig:compare_order}
\end{figure*}

\section{Experiment}

\subsection{Implementation Details}

\subsubsection{Dataset and Benchmark.}
We train our model on a high-quality, self-curated video dataset (see Section 3.2). Starting from 15M clips obtained after scene detection, we perform a two-stage filtering process to ensure semantic alignment and appropriate subject visibility. This yields 2.5M high-quality clips, each paired with multiple reference images stored as RGBA. Each clip consists of 81 frames. From these, we select 1M unique clips to form the final training set MoFu-1M.
To evaluate scale consistency and permutation invariance, we construct MoFu-Bench, a dedicated benchmark with 1,000 subject-text pairs. Cases are partially adapted from ConsisID~\cite{consisid}, A2-Bench~\cite{skyreels}, and OpenS2V-Bench~\cite{opens2v}, with the remainder curated for broad subject diversity. Each case includes up to three reference images with systematic scale variations and randomized permutations, paired with a high-quality prompt for semantic alignment.  
Compared to existing benchmarks, MoFu-Bench is the first to explicitly assess both scale consistency and permutation invariance, providing diverse subjects and fine-grained annotations as a strong standard for future evaluation.

\subsubsection{Evaluation Metrics.}
To comprehensively evaluate generated videos, upon MoFu-Bench, we adopt six metrics covering visual quality, subject fidelity, temporal stability, and scale consistency: 
Aesthetics~\cite{aes}, FaceSim~\cite{face}, GmeScore, Motion Score, ScaleScore, and SubjectSim. 
Aesthetics evaluates perceptual quality, FaceSim and SubjectSim measure identity preservation, GmeScore quantifies text-video alignment, Motion Score assesses temporal coherence, and ScaleScore evaluates relative subject scales against prompt descriptions. 
Detailed definitions and implementations are provided in the Appendix.

\begin{table*}[h]
\centering
\small
\label{tab:quantitative_comparison}
\begin{tabular}{c|c c c c c c}
\toprule[1.1pt]
\textbf{Method} & 
\textbf{Aesthetics$\uparrow$} & 
\textbf{FaceSim$\uparrow$} &
\textbf{GmeScore$\uparrow$} & 
\textbf{Motion$\leftrightarrow$} & 
\textbf{ScaleScore$\uparrow$} & 
\textbf{SubjectSim$\uparrow$} \\
\midrule
\multirow{4}{*}{}
Phantom~\cite{phantom}     & 0.355 & \underline{0.375} & 0.706 & 0.229 & 0.536 & \underline{0.748} \\
SkyReels-A2~\cite{skyreels} & 0.286 & 0.341 & 0.691 & 0.233 & 0.527 & 0.737 \\
VACE~\cite{vace}        & \underline{0.392} & 0.247 & \underline{0.732} & 0.214 & \underline{0.547} & 0.692 \\
MAGREF~\cite{magref}      & 0.369 & 0.362 & 0.717 & 0.207 & 0.511 & 0.731 \\
\midrule
MoFu & \textbf{0.401} & \textbf{0.396} & \textbf{0.745} & 0.221 & \textbf{0.585} & \textbf{0.755} \\
\bottomrule
\end{tabular}
\caption{Quantitative comparison on MoFu-Bench. The best scores are \textbf{bolded}, while the second-best is \underline{underlined}.}
\label{tab:quantitative_comparison}
\end{table*}
\begin{figure*}[htbp]
    \centering
    \includegraphics[width=\textwidth]{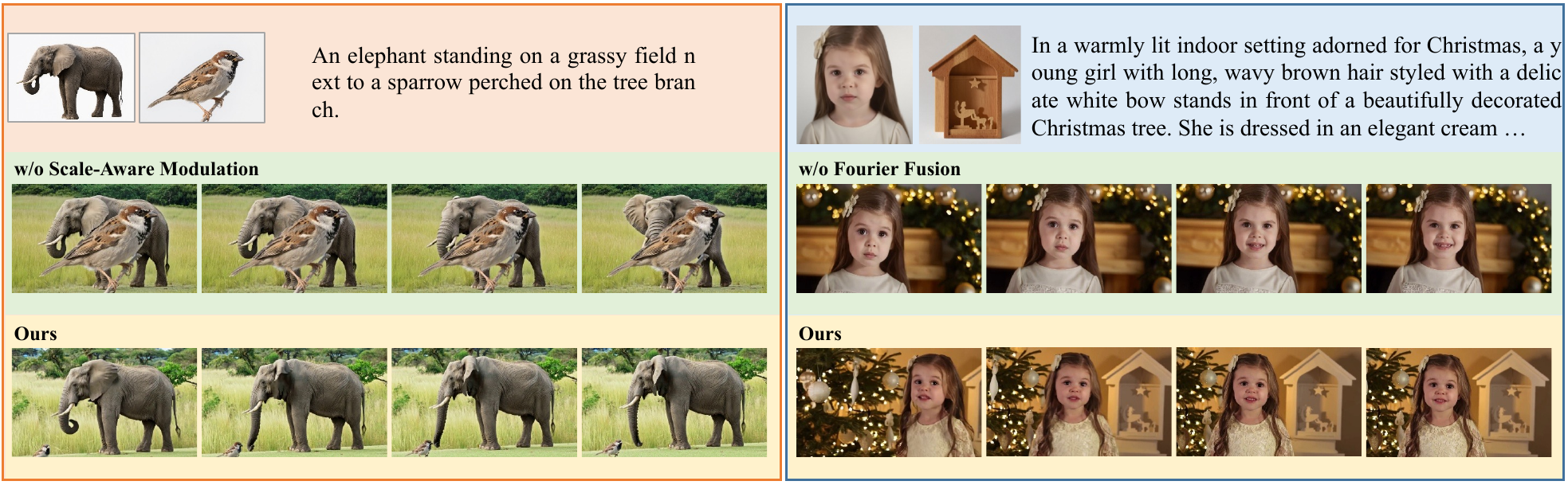}
    \caption{Ablation studies on MoFu.
\textbf{Left:} Without SMO, the relative scale between the elephant and sparrow becomes unrealistic. With SMO, the natural size relationship is preserved.
\textbf{Right:} Without Fourier Fusion, the model is sensitive to the order and may drop subjects. With Fourier Fusion, MoFu consistently generates all subjects regardless of reference permutations.}
    \label{fig:ablation}
\end{figure*}

\subsubsection{Training Strategies.}
We train our model using the AdamW optimizer, configured with $\beta_1 = 0.9$, $\beta_2 = 0.999$, and a weight decay of 0.01. The learning rate is initialized at $1 \times 10^{-5}$ and follows a cosine annealing schedule with periodic restarts. Model training is conducted on 16 NVIDIA H800 GPUs for 7 days. Input videos are processed at a resolution of 480P, with a sequence length of 81 frames.

\subsection{Qualitative Analysis}

Fig.~\ref{fig:qualitative} presents qualitative results of MoFu, showing its ability to generate high-quality videos that align with prompts while maintaining natural scales and stability across reference permutations. 
In single- and multi-subject cases (1–4), MoFu preserves subject identity and consistent scales even under challenging settings, while Case 5 demonstrates its robustness to scale discrepancies in reference images. 
As shown in Fig.~\ref{fig:compare_scale} and Fig.~\ref{fig:compare_order}, MoFu surpasses prior methods in both scale-inconsistent and permutation-variant scenarios, accurately preserving relative object sizes and spatial relations (e.g., bowl–truck, cup–car) with realistic motion patterns. 
Overall, MoFu achieves prompt-faithful, scale-consistent, and permutation-invariant video generation across diverse multi-subject scenes.


\subsection{Quantitative Results}
As shown in Tab.~\ref{tab:quantitative_comparison}, MoFu achieves state-of-the-art performance across nearly all metrics on MoFu-Bench.
It attains the highest Aesthetics and FaceSim scores, producing visually appealing videos with strong identity preservation.
Although I2V-based methods such as MAGREF perform well on FaceSim and SubjectSim due to strong reference conditioning, their scale inconsistency limits overall quality.
MoFu also leads on GmeScore, reflecting superior text–video alignment, and achieves balanced Motion results by maintaining temporal coherence without sacrificing fidelity.
Moreover, it yields substantial gains on ScaleScore and SubjectSim, demonstrating robust scale consistency and subject stability across permutations.

\subsection{Ablation studies}
\textbf{Effect of SMO and Fourier Fusion.}  
We ablate SMO and Fourier Fusion to assess their contributions as shown in Fig.~\ref{fig:ablation}. Without SMO, the model fails to maintain natural relative scales. For example, the sparrow appears disproportionately large compared to the elephant. Incorporating SMO explicitly encodes scale cues from the prompt, preserving realistic proportions.  
Removing Fourier Fusion makes the model sensitive to reference order, causing missing subjects, like the wooden box in the case, under permutation. Fourier Fusion eliminates this sensitivity, producing consistent and complete outputs regardless of permutations.  

\textbf{Scale-Permutation Stability Loss.}  
SPSL consists of a scale loss and a permutation loss. We do not provide ablations of these two components in Fig.~\ref{fig:ablation} because SMO and FF strategy are supervised respectively by the scale and permutation information provided by this loss. Removing the loss while training SMO or FF independently would render the comparison meaningless, as these modules would lack the necessary supervision signals. Therefore, we omit ablations on the Scale-Permutation Stability Loss.  

\section{Conclusion}
We proposed MoFu, a unified framework that addresses scale inconsistency and permutation sensitivity in multi-subject video generation. By introducing Scale-Aware Modulation and the Fourier Fusion strategy, MoFu explicitly enforces natural subject scales and permutation-invariant conditioning. 
Besides, we build MoFu-1M for training, and establish a dedicated benchmark MoFu-Bench for assessing scale consistency and permutation-invariance.

\section*{Acknowledgments}
This work is partially supported by the National Natural Science Foundation of China under Grant No. 62576083.

\bibliography{aaai2026}
\clearpage

\appendix

\maketitlesupplementary

\section{Benchmark}
\subsection{Distribution of MoFu-Bench}

\begin{figure}[t]   
    \centering
    \includegraphics[width=\linewidth,scale=1.00]{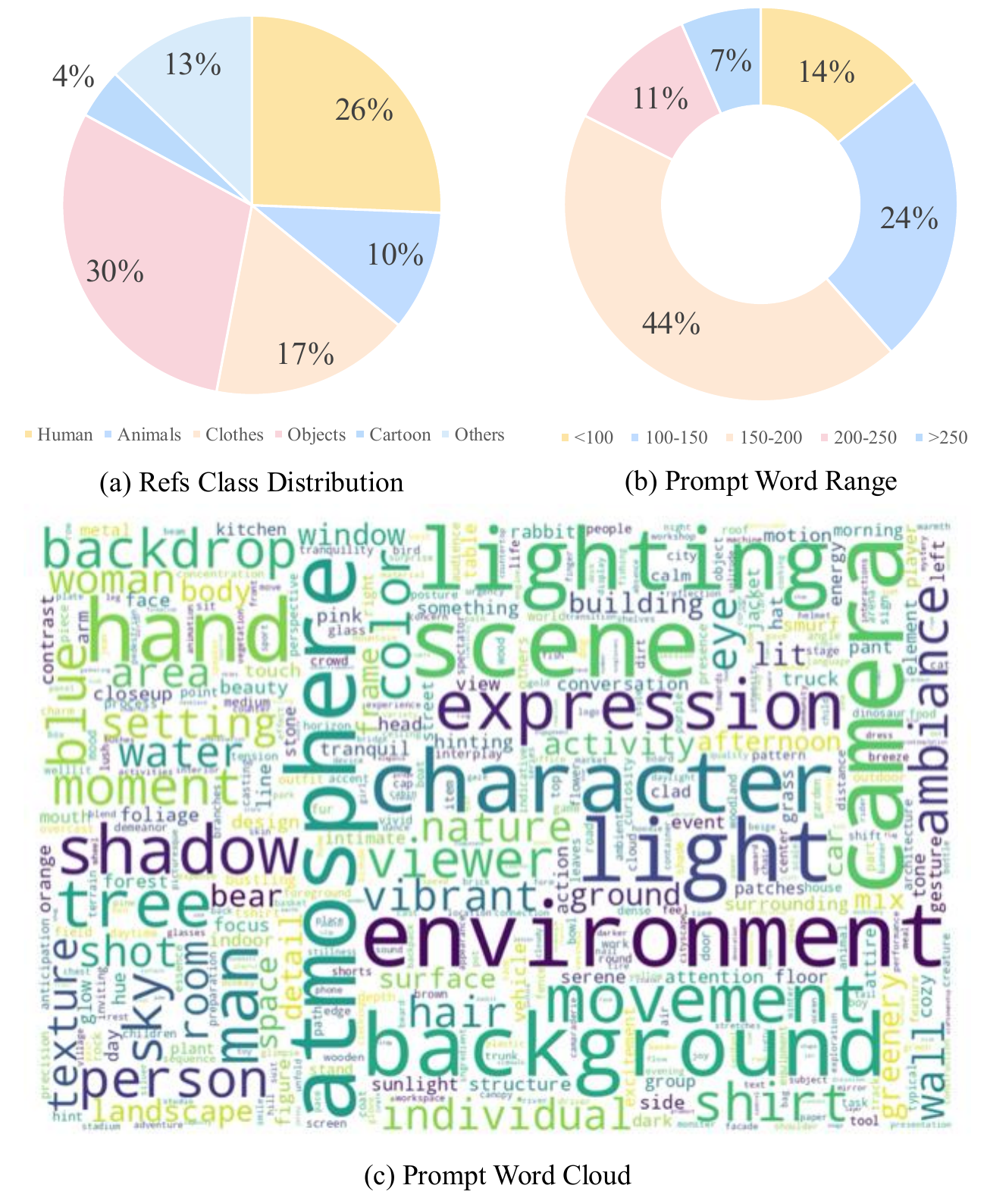}
    \caption{MoFu-Bench Statistics.}
    \label{fig:benchmark_stat}
\end{figure}

To better understand the composition and diversity of our benchmark, we present a detailed analysis of MoFu-Bench in Fig.~\ref{fig:benchmark_stat}.  
As shown in Fig.~\ref{fig:benchmark_stat}(a), the reference images in MoFu-Bench span a wide range of categories, including \textit{human} (26\%), \textit{animals} (17\%), \textit{clothing} (10\%), \textit{objects} (30\%), \textit{cartoons} (4\%), and \textit{others} (13\%). This broad coverage ensures that the benchmark evaluates models across both realistic and stylized domains.
Fig.~\ref{fig:benchmark_stat}(b) shows the distribution of prompt lengths, with most prompts falling between 100 and 200 words. These relatively long and descriptive prompts encourage models to generate videos with rich semantics and nuanced subject interactions.
Additionally, Fig.~\ref{fig:benchmark_stat}(c) presents a word cloud of the most frequently used words in the prompts. Terms related to actions, environments, and subjects frequently appear, highlighting the benchmark’s focus on multi-subject interaction in diverse real-world settings.
Overall, the balanced subject distribution, semantically rich prompts, and wide-ranging domains make MoFu-Bench a comprehensive testbed for evaluating scale consistency and permutation invariance in multi-subject video generation.

\section{Prompt used in MoFu}

\begin{figure}[t]   
    \centering
    \includegraphics[width=\linewidth,scale=1.00]{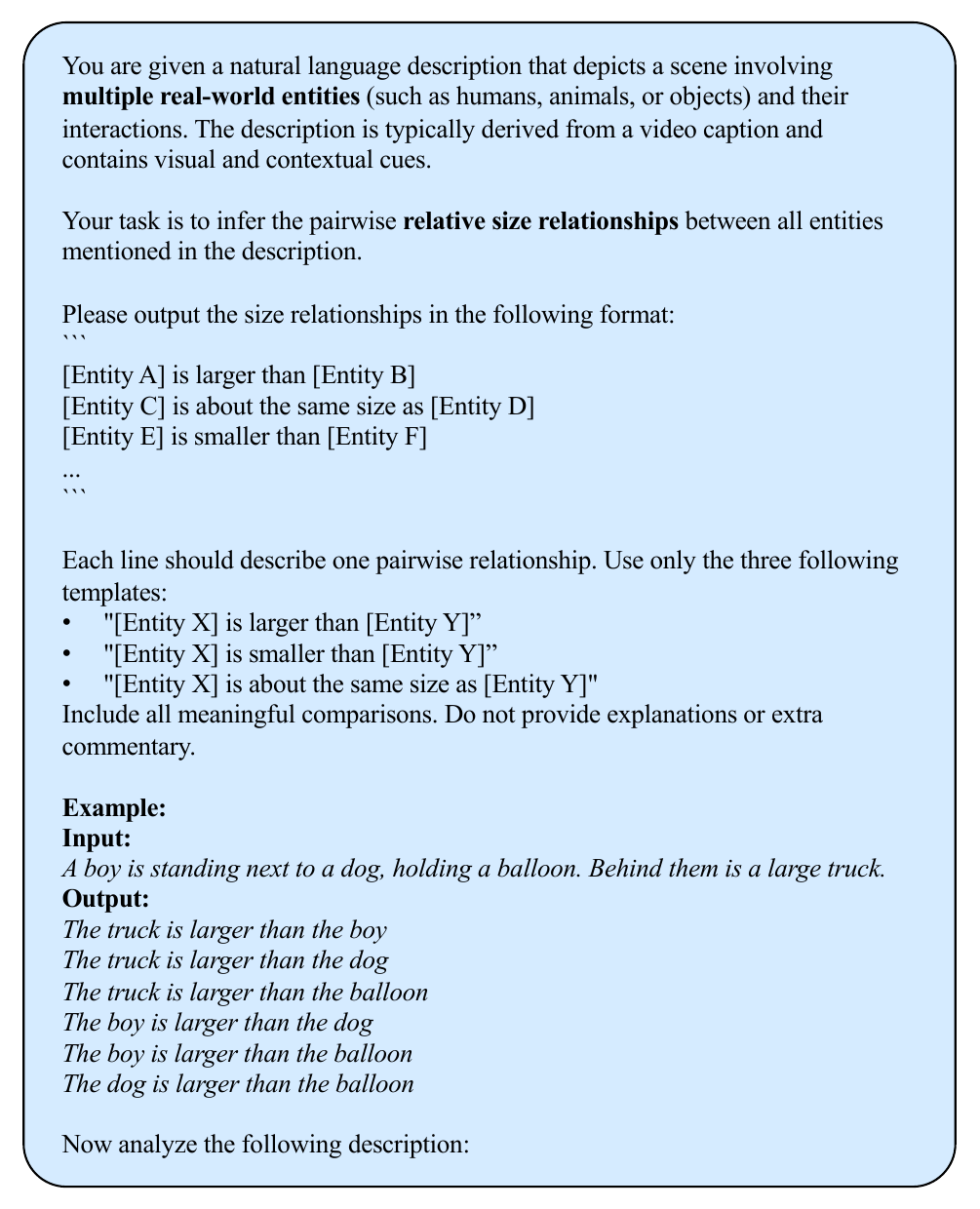}
    \caption{Prompt for MoFu to extract scale relationships.}
    \label{fig:prompt_mofu}
\end{figure}

To extract scale-aware semantics from textual prompts, we design an LLM instruction that guides the model~\cite{qwen2.5} to infer pairwise relative size relationships among entities mentioned in the scene description. As shown in Fig.~\ref{fig:prompt_mofu}, the prompt provides structured templates and avoids free-form generation to ensure stable and comparable outputs.

Rather than using the textual output itself, we extract the \texttt{[CLS]} embedding from the LLM as the final scale-aware representation. This embedding is the input to Scale Control Adapter and then used to guide the Scale-Aware Modulation module during video generation.

\begin{figure}[t]   
    \centering
    \includegraphics[width=\linewidth,scale=1.00]{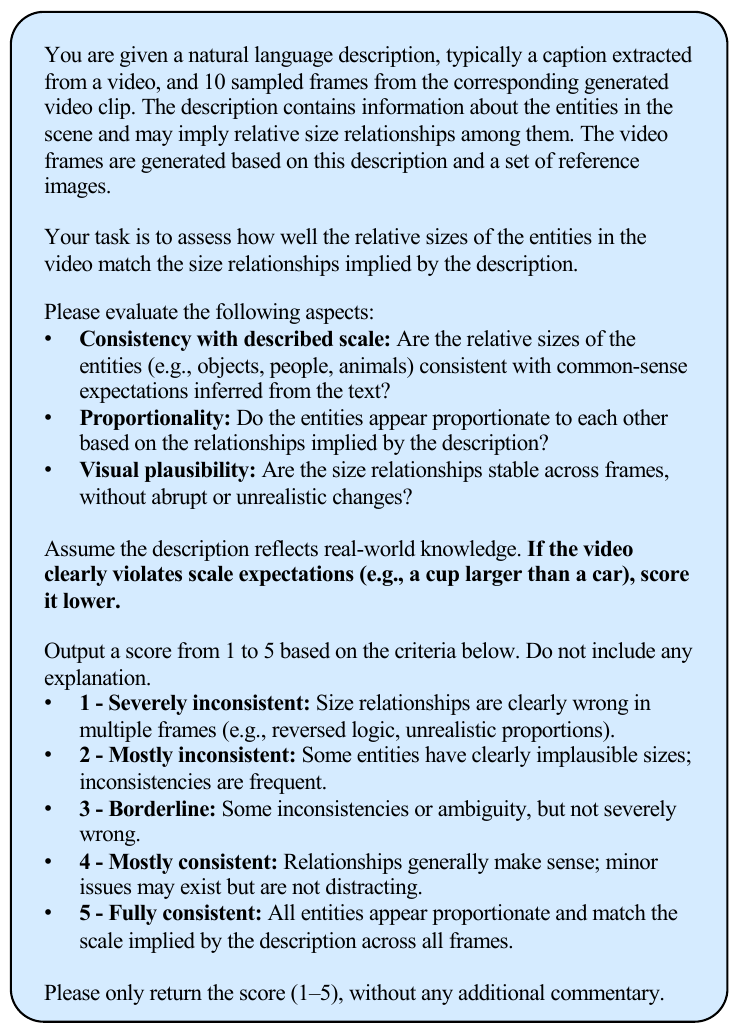}
    \caption{Prompt for GPT to calculate ScaleScore.}
    \label{fig:prompt_scalescore}
\end{figure}

\begin{figure*}[h]   
    \centering
    \includegraphics[width=\textwidth,scale=1.00]{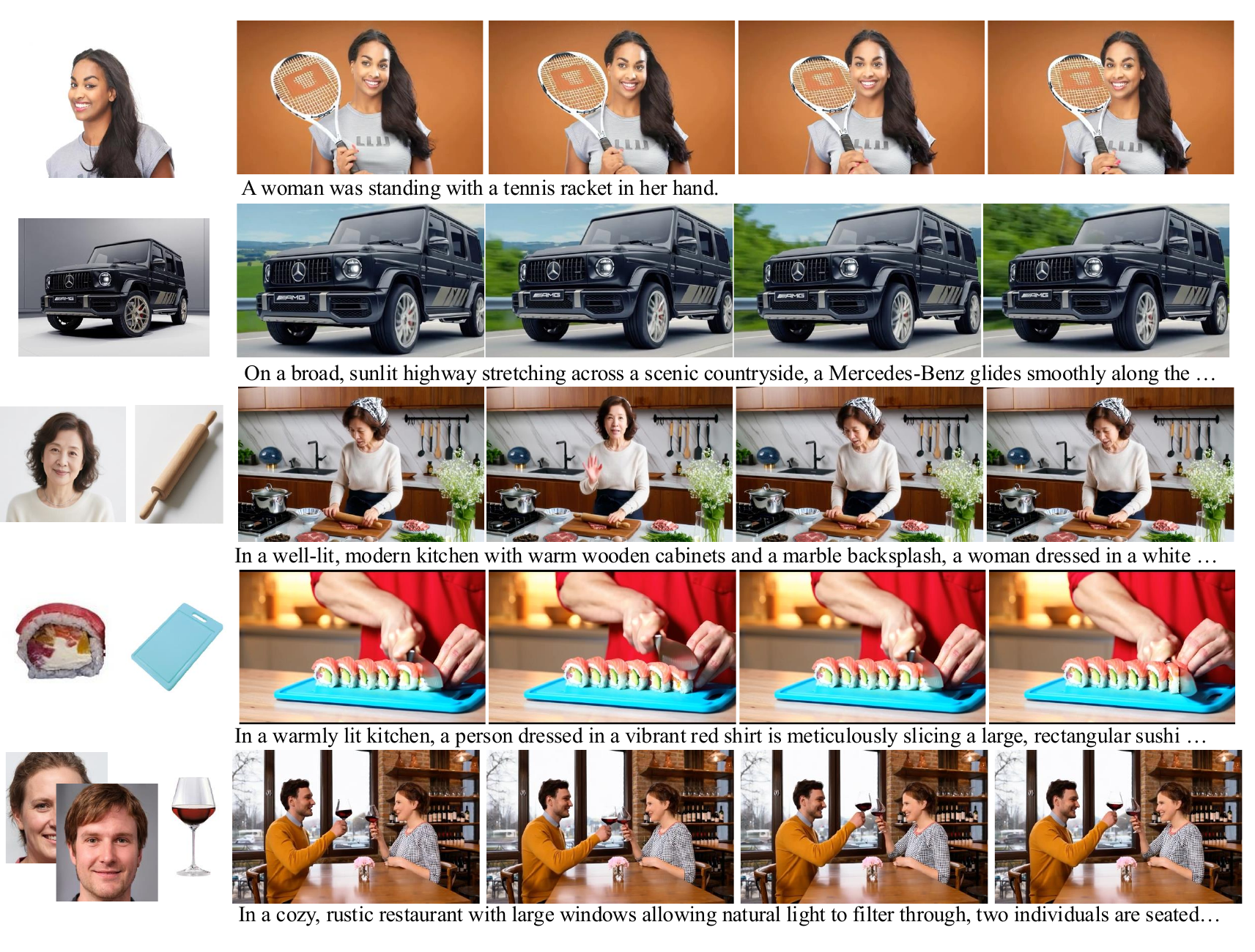}
    \caption{Cases by MoFu.}
    \label{fig:appendix_case_study}
\end{figure*}

\section{Experiments}
\subsection{Evaluation Metrics.}

Following the evaluation principles established in OpenS2V~\cite{opens2v}, we design a comprehensive metric suite that assesses visual quality, identity fidelity, motion stability, and scale consistency in multi-subject video generation. Specifically, we adopt six complementary metrics:

\begin{itemize}
    \item \textbf{Aesthetics:}  
    We evaluate the perceptual quality and artistic appeal of the generated frames with Aesthetics Score~\cite{aes}, using an aesthetics prediction model trained on large-scale image quality datasets. Higher scores indicate visually pleasing and artifact-free results.  

    \item \textbf{FaceSim:}  
    For videos containing human subjects, we employ FaceSim~\cite{face} to measure the similarity between generated faces and reference faces using a face recognition network (e.g., ArcFace). This ensures that identity-specific details such as facial structure and attributes are preserved.  

    \item \textbf{GmeScore:}  
    We use the GmeScore, a retrieval-based metric built on a Qwen2-VL~\cite{qwen2vl} model fine-tuned for vision-language alignment. This metric quantifies how accurately the model maintains subject shapes and key structural features.

    \item \textbf{Motion Score:} 
    We assess temporal coherence by Motion Score, implemented by OpenCV~\cite{opencv}, measuring inter-frame motion stability using optical flow smoothness and frame-level warping error. Lower motion artifacts yield higher scores.  

    \item \textbf{ScaleScore:}  
    To evaluate ScaleScore, we leverage GPT-4o~\cite{gpt4o} to judge whether the relative sizes of subjects in the generated video match the relationships implied by the input prompt. We provide GPT-4o with the text description and 10 sampled frames from the video, and ask it to output a scale consistency score from 1 to 5. The prompt is illustrated in the Fig.~\ref{fig:prompt_scalescore}.

    \item \textbf{SubjectSim:}  
    SubjectSim measures the overall similarity between generated subjects and their reference images, combining accurate subject localization and feature similarity.  
    First, we obtain subject masks in reference images using GroundSAM~\cite{groundedsam}. For sampled 10 generated frames, candidate subject regions are detected with the same detector and matched to the reference subject via CLIP~\cite{clip} feature similarity:
    $$
    j^{*} = \arg\max_j \frac{f(b_{t,j}) \cdot f(S^{ref}_i)}{\|f(b_{t,j})\| \|f(S^{ref}_i)\|},
    $$
    where \(b_{t,j}\) is the \(j\)-th detected region at frame \(t\) and \(S^{ref}_i\) is the reference subject.  
    We then compute the cosine similarity of the matched regions' CLIP embeddings and average over all frames and subjects:
    \[
    \text{SubjectSim} = \frac{1}{N} \sum_{i=1}^{N} \frac{1}{T} \sum_{t=1}^{T} 
    \frac{f(S^{gen}_{i,t}) \cdot f(S^{ref}_i)}
    {\|f(S^{gen}_{i,t})\| \|f(S^{ref}_i)\|}.
    \]
    Higher SubjectSim values indicate better preservation of subject identity and appearance.
\end{itemize}

Together, these metrics provide a holistic evaluation of spatial coherence, subject identity preservation, temporal stability, and scale robustness.




\subsection{Case Study}
To further demonstrate the robustness and versatility of MoFu, we provide additional case studies covering a broader range of scenes, subjects, and challenging conditions. The complete set of extended qualitative examples is shown in Fig.~\ref{fig:appendix_case_study}, where each case includes the input prompt, reference images, and representative frames from the generated videos. 


\section{Limitations}
While MoFu demonstrates strong scale consistency and permutation invariance, its design inherently emphasizes robust spatial reasoning, which may limit its flexibility in modeling extremely fine-grained temporal interactions in highly dynamic scenes. The framework also benefits from high-quality reference inputs, and its performance could degrade when references are sparse or noisy, though this strict reliance may also encourage more faithful subject reconstruction. Additionally, MoFu adopts a unified architecture tailored for general multi-subject generation; while this ensures broad applicability, it may overlook domain-specific priors that could further enhance performance.
\end{document}